\documentclass[10pt,twocolumn,letterpaper]{article}

\usepackage[pagenumbers]{cvpr}   

\usepackage[accsupp]{axessibility}
\usepackage{amsmath}
\usepackage{amssymb}
\usepackage{booktabs}
\usepackage[misc]{ifsym}
\usepackage{graphicx}
\usepackage{multirow}
\usepackage{pifont}
\usepackage{tabularx}

\newcommand{\cmark}{\textcolor{green}{\ding{51}}}%
\newcommand{\xmark}{\textcolor{red}{\ding{55}}}%

\usepackage{tikz}
\usepackage{pgf, pgfsys, pgffor}
\usepackage{pgfplotstable}
\pgfplotsset{compat=1.15}

\usepackage[pagebackref=true,breaklinks=true,colorlinks,citecolor=blue,linkcolor=red,bookmarks=false]{hyperref}

\usepackage[capitalize]{cleveref}
\crefname{section}{Sec.}{Secs.}
\Crefname{section}{Section}{Sections}
\Crefname{table}{Table}{Tables}
\crefname{table}{Tab.}{Tabs.}

\def\paperID{233} 
\def\confName{CVPR}
\def\confYear{2024}

\begin{document}

\newcolumntype{Y}{>{\centering\arraybackslash}X}

\title{CityDreamer: Compositional Generative Model of Unbounded 3D Cities}

\author{%
Haozhe Xie, Zhaoxi Chen, Fangzhou Hong, Ziwei Liu
\textsuperscript{\Letter} \\
S-Lab, Nanyang Technological University\\
{%
\tt\small \{haozhe.xie, zhaoxi001, fangzhou001, ziwei.liu\}@ntu.edu.sg}\\
\tt\small \url{https://haozhexie.com/project/city-dreamer}
}

\twocolumn[{%
\renewcommand\twocolumn[1][]{#1}%
\maketitle
\begin{center}
  \vspace{-9 mm}
  \centering
  \captionsetup{type=figure}
  \includegraphics[width=\textwidth]{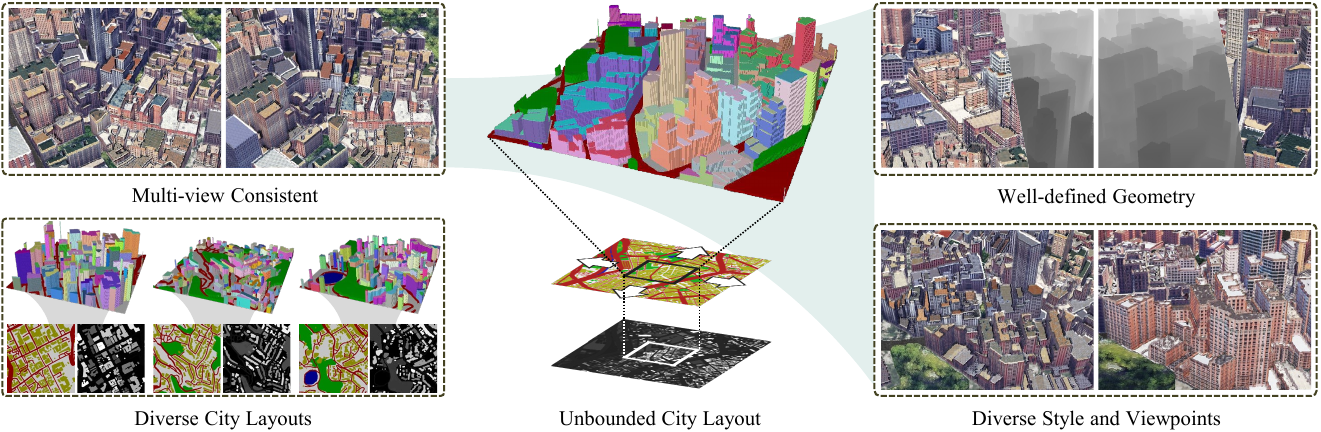}
  \vspace{-6.5 mm}
  \captionof{figure}{The proposed CityDreamer generates a wide variety of unbounded city layouts and multi-view consistent appearances, featuring well-defined geometries and diverse styles.}
\end{center}%
}]


\begin{abstract}
\vspace{-4 mm}
3D city generation is a desirable yet challenging task, since humans are more sensitive to structural distortions in urban environments.
Additionally, generating 3D cities is more complex than 3D natural scenes since buildings, as objects of the same class, exhibit a wider range of appearances compared to the relatively consistent appearance of objects like trees in natural scenes.
To address these challenges, we propose \textbf{CityDreamer}, a compositional generative model designed specifically for unbounded 3D cities.
Our key insight is that 3D city generation should be a composition of different types of neural fields: \textbf{1)} various building instances, and \textbf{2)} background stuff, such as roads and green lands.
Specifically, we adopt the bird's eye view scene representation and employ a volumetric render for both \textbf{instance-oriented} and \textbf{stuff-oriented} neural fields.
The generative hash grid and periodic positional embedding are tailored as scene parameterization to suit the distinct characteristics of building instances and background stuff.
Furthermore, we contribute a suite of \textbf{CityGen Datasets}, including OSM and GoogleEarth, which comprises a vast amount of real-world city imagery to enhance the realism of the generated 3D cities both in their layouts and appearances.
CityDreamer achieves state-of-the-art performance not only in generating realistic 3D cities but also in localized editing within the generated cities.
\vspace{-5 mm}
\end{abstract}

\vspace{-12 mm}
\section{Introduction}

In the wave of the metaverse, 3D asset generation has drawn considerable interest. 
Significant advancements have been achieved in generating 3D objects~\cite{DBLP:conf/iccv/XieYSZZ19,DBLP:journals/ijcv/XieYZZS20,DBLP:arxiv/2303-13508}, 3D avatars~\cite{DBLP:conf/iclr/Hong0LP023,DBLP:arxiv/2305-19012,DBLP:arxiv/2306-09329}, and 3D scenes~\cite{DBLP:conf/cvpr/ChaiTLIS23,DBLP:arxiv/2302-01330,DBLP:conf/iccv/LinLMCSYT23}.
Cities, being one of the most crucial 3D assets, have found widespread use in various applications, including urban planning, environmental simulations, and game asset creation.
Therefore, the quest to make 3D city development accessible to a broader audience encompassing artists, researchers, and players, becomes a significant and impactful challenge.

In recent years, notable advancements have been made in the field of 3D scene generation. 
GANCraft~\cite{DBLP:conf/iccv/HaoMB021} and SceneDreamer~\cite{DBLP:arxiv/2302-01330} use volumetric neural rendering to produce images within the 3D scene, using 3D coordinates and corresponding semantic labels. 
Both methods show promising results in generating 3D natural scenes by leveraging pseudo-ground-truth images generated by SPADE~\cite{DBLP:conf/cvpr/Park0WZ19}.
A very recent work, InfiniCity~\cite{DBLP:conf/iccv/LinLMCSYT23}, follows a similar pipeline for 3D city generation. 
However, creating 3D cities presents greater complexity compared to 3D natural scenes. 
Buildings, as objects with the same semantic label, exhibit a wide range of appearances, unlike the relatively consistent appearance of objects like trees in natural scenes.
This fact may decrease the quality of generated buildings when all buildings in a city are given the same semantic label.

To handle the diversity of buildings in urban environments, we propose CityDreamer, a compositional generative model designed for unbounded 3D cities.
As shown in Figure~\ref{fig:overview}, CityDreamer differs from existing methods in that it splits the generation of building instances and background stuff like roads, green lands, and water areas into two separate modules: the building instance generator and the city background generator.
Both generators adopt the bird's eye view (BEV) scene representation and employ a volumetric renderer to generate photorealistic images via adversarial training.
Notably, the scene parameterization is meticulously tailored to suit the distinct characteristics of background stuff and buildings.
Background stuff in each category typically has similar appearances while exhibiting irregular textures. 
Hence, we introduce the generative hash grid to preserve naturalness while upholding 3D consistency.
In contrast, building instances exhibit a wide range of appearances, but the texture of their fa\c{c}ades often displays regular periodic patterns.
Therefore, we design periodic positional encoding, which is simple yet effective for handling the diversity building fa\c{c}ades.
The compositor finally combines the rendered background stuff and building instances to generate a cohesive image.

To enhance the realism of our generated 3D cities, we construct a suite of CityGen Datasets, including OSM and GoogleEarth.
The OSM dataset, sourced from OpenStreetMap~\cite{HAOZHE:link/OpenStreetMap}, contains semantic maps and height fields of 80 cities, covering over 6,000 km$^2$. 
These maps show the locations of roads, buildings, green lands, and water areas, while the height fields primarily indicate building heights.
The GoogleEarth dataset, gathered using Google Earth Studio~\cite{HAOZHE:link/GoogleEarth}, features 400 orbit trajectories in New York City.
It includes 24,000 real-world city images, along with semantic and building instance segmentation. These annotations are automatically generated by projecting the 3D city layout, based on the OSM dataset, onto the images.
The Google Earth dataset provides a wider variety of realistic urban images from different perspectives. Additionally, it can be easily expanded to include cities worldwide.

The contributions are summarized as follows:

\begin{itemize}
\setlength\itemsep{0 em}
\item We propose CityDreamer, a compositional generative model designed specifically for unbounded 3D cities, which disentangles instance-oriented and stuff-oriented neural fields for buildings and backgrounds.
\item We construct the CityGen Datasets, including OSM and GoogleEarth, with realistic city layouts and appearances, respectively. GoogleEarth includes images with multi-view consistency and building instance segmentation.
\item The proposed CityDreamer showcases its superior capability in generating large-scale and diverse 3D cities. Additionally, it enables localized editing within the generated cities.
\end{itemize}

\section{Related Work}

\noindent \textbf{3D-Aware GANs.}
Generative adversarial networks (GANs)~\cite{DBLP:conf/nips/GoodfellowPMXWOCB14} have achieved remarkable success in 2D image generation~\cite{DBLP:journals/pami/KarrasLA21,DBLP:conf/cvpr/KarrasLAHLA20}. 
Efforts to extend GANs into 3D space have also emerged, with some works~\cite{DBLP:conf/iccv/Nguyen-PhuocLTR19,DBLP:conf/3dim/GadelhaMW17,DBLP:conf/nips/0001ZXFT16} intuitively adopting voxel-based representations by extending the CNN backbone used in 2D. 
However, the high computational and memory cost of voxel grids and 3D convolution poses challenges in modeling unbounded 3D scenes.
Recent advancements in neural radiance field (NeRF)~\cite{DBLP:conf/eccv/MildenhallSTBRN20} have led to the incorporation of volume rendering as a key inductive bias to make GANs 3D-aware. 
This enables GANs to learn 3D representations from 2D images~\cite{DBLP:conf/cvpr/ChanLCNPMGGTKKW22,DBLP:conf/iclr/GuL0T22,DBLP:conf/cvpr/Or-ElLSSPK22,DBLP:conf/cvpr/XueLSL22,DBLP:conf/nips/0004SWCYLLGF22}. 
Nevertheless, most of these methods are trained on curated datasets for bounded scenes, such as human faces~\cite{DBLP:journals/pami/KarrasLA21}, human bodies~\cite{DBLP:journals/pami/IonescuPOS14}, and objects~\cite{DBLP:conf/cvpr/WuZFWRPWYWQLL23}.

\noindent \textbf{Scene-Level Content Generation.}
Unlike impressive 2D generative models that mainly target single categories or common objects, generating scene-level content is a challenging task due to the high diversity of scenes. 
Semantic image synthesis, such as~\cite{DBLP:conf/iccv/HaoMB021,DBLP:conf/cvpr/Park0WZ19,DBLP:conf/cvpr/EsserRO21,DBLP:conf/eccv/MallyaWS020}, shows promising results in generating scene-level content in the wild by conditioning on pixel-wise dense correspondence, such as semantic segmentation maps or depth maps. 
Some approaches have even achieved 3D-aware scene synthesis~\cite{DBLP:conf/iccv/HaoMB021,DBLP:conf/iccv/LiuM0SJK21,DBLP:conf/eccv/LiWSK22,DBLP:conf/eccv/MallyaWS020,DBLP:conf/eccv/ShiSZYC22}, but they may lack full 3D consistency or support feed-forward generation for novel worlds. 
Recent works like~\cite{DBLP:conf/cvpr/ChaiTLIS23,DBLP:conf/iccv/LinLMCSYT23,DBLP:arxiv/2302-01330} have achieved 3D consistent scenes at infinity scale through unbounded layout extrapolation.
Another bunch of work~\cite{DBLP:conf/iccv/DeVries0STS21,DBLP:conf/nips/PaschalidouKSKG21,DBLP:conf/3dim/WangYN21,DBLP:conf/nips/0001GATTCDZGUDS22}  focus on indoor scene synthesis using expensive 3D datasets~\cite{DBLP:conf/cvpr/DaiCSHFN17,DBLP:arxiv/1906-05797} or CAD retrieval~\cite{DBLP:conf/iccv/FuC0ZWLZSJZ021}.

\section{Our Approach}

\begin{figure*}[!t]
  \centering
  \includegraphics[width=\linewidth]{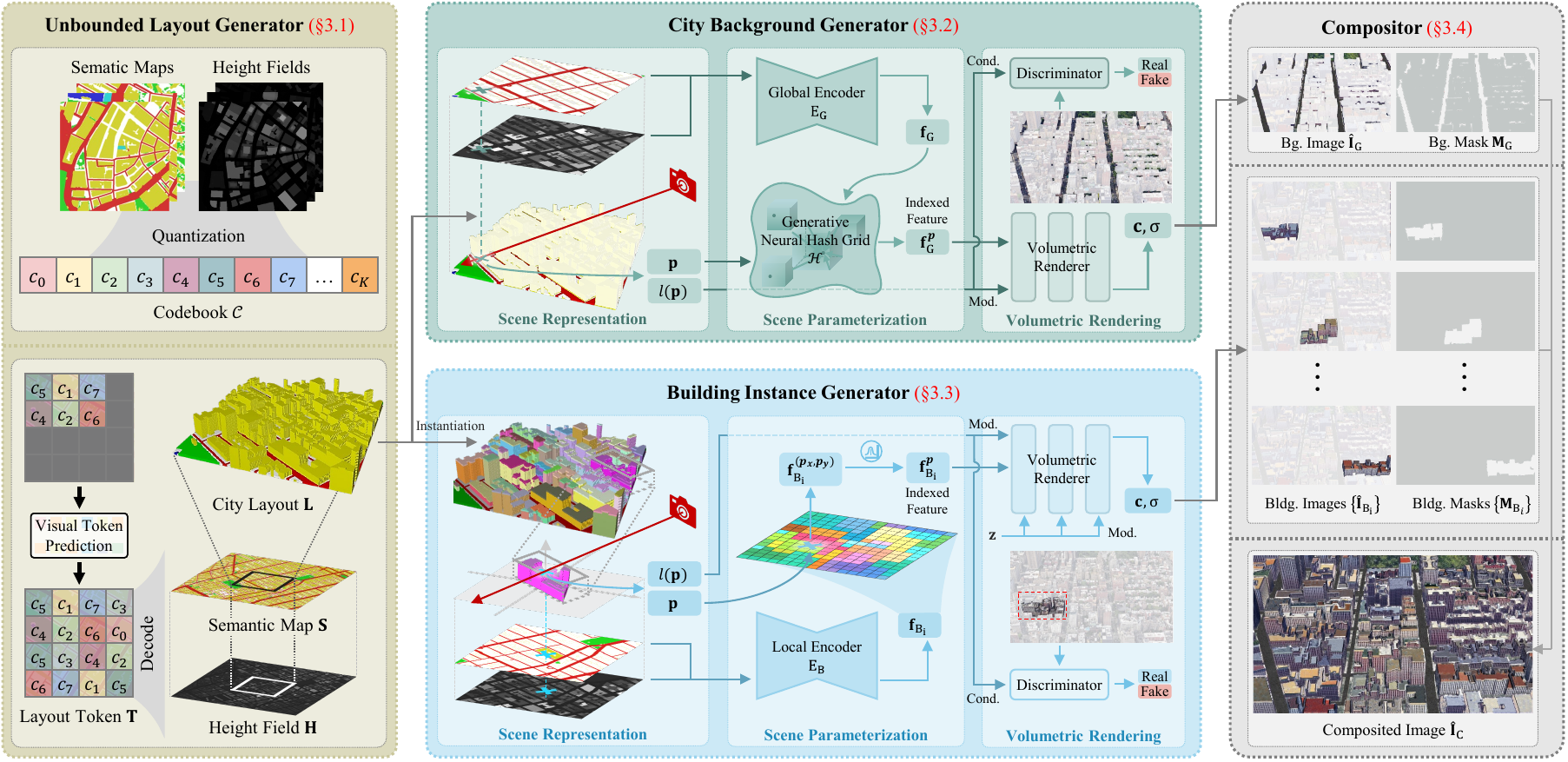}
  \caption{\textbf{Overview of CityDreamer.}
  The \textit{unbounded layout generator} creates the city layout $\mathbf{L}$. 
  Then, the \textit{city background generator} performs ray-sampling to retrieve features from $\mathbf{L}$ and generates the background image with a volumetric renderer focusing on background stuff like roads, green lands, and water areas. 
  Similarly, the \textit{building instance generator} renders the building instance image with another volumetric renderer. 
  Finally, the \textit{compositor} merges the rendered background and building images, producing a unified and coherent final image. 
  Note that ``Mod.'', ``Cond.'', ``Bg.'', and ``Bldg.'' denote ``Modulation'', ``Condition'', ``Background'', and ``Building'', respectively.}
  \label{fig:overview}
\end{figure*}

As shown in Figure~\ref{fig:overview}, CityDreamer follows a four-step process to generate an unbounded 3D city. 
Initially, the unbounded layout generator (Sec.~\ref{sec:unbounded-city-layout-generator}) creates an arbitrary large city layout $\mathbf{L}$. 
Subsequently, the city background generator (Sec.~\ref{sec:city-bg-generator}) produces the background image $\mathbf{\hat{I}}_{\rm G}$ along with its corresponding mask $\mathbf{M}_{\rm G}$. 
Next, the building instances generator (Sec.~\ref{sec:building-ins-generator}) generates images for building instances $\{\mathbf{\hat{I}}_{\rm B}^i\}_{i=1}^n$ and their respective masks $\{\mathbf{M}_{\rm B}^i\}_{i=1}^n$, where $n$ is the number of building instances.
Lastly, the compositor (Sec.~\ref{sec:compositor}) merges the rendered background and building instances into a single cohesive image $\mathbf{I}_{\rm C}$.

\subsection{Unbounded City Layout Generator}
\label{sec:unbounded-city-layout-generator}

\noindent \textbf{City Layout Represenetation.}
The city layout determines the 3D objects present in the city and their respective locations.
The objects can be categorized into six classes: roads, buildings, green lands, construction sites, water areas, and others.
Moreover, there is an additional null class used to represent empty spaces in the 3D volumes.
The city layout in CityDreamer, denoted as a 3D volume $\mathbf{L}$, is created by extruding the pixels in the semantic map $\mathbf{S}$ based on the corresponding values in the height field $\mathbf{H}$.
Specifically, the value of $\mathbf{L}$ at $(i, j, k)$ can be defined as

\begin{equation}
  L_{(i, j, k)} = 
  \begin{cases}
    S_{(i, j)} & {\rm if}~k \leq H_{(i, j)} \\
    0          & {\rm otherwise}
  \end{cases}
  \label{eq:layout-gen}
\end{equation}
where $0$ denotes empty spaces in the 3D volumes.

\noindent \textbf{City Layout Generation.}
Obtaining unbounded city layouts is translated into generating extendable semantic maps and height fields.
To this aim, we construct the unbounded layout generator based on MaskGIT~\cite{DBLP:conf/cvpr/ChangZJLF22}, which inherently enables inpainting and extrapolation capabilities.
Specifically, we employ VQVAE~\cite{DBLP:conf/nips/RazaviOV19,DBLP:conf/nips/OordVK17} to tokenize the semantic map and height field patches, converting them into discrete latent space and creating a codebook $\mathcal{C} = \{c_k | c_k \in \mathbb{R}^{D}\}_{i=1}^K$.
During inference,  we generate the layout token $\mathbf{T}$ in an autoregressive manner, and subsequently, we use the VQVAE's decoder to generate a pair of semantic map $\mathbf{S}$ and height field $\mathbf{H}$.
Since VQVAE generates fixed-size semantic maps and height fields, we use image extrapolation to create arbitrary-sized ones.
During this process, we adopt a sliding window to forecast a local layout token at every step, with a 25\% overlap during the sliding.

\noindent \textbf{Loss Functions.}
The VQVAE treats the generation of the height field and semantic map as two separate tasks, optimizing them using L1 Loss and Cross Entropy Loss $\mathcal{E}$, respectively. 
Additionally, to ensure sharpness in the height field around the edges of the buildings, we introduce an extra Smoothness Loss $\mathcal{S}$~\cite{DBLP:conf/aaai/MeisterH018}.
\begin{equation}
  \ell_{\rm VQ} = \lambda_{\rm R} \lVert\mathbf{\hat{H}_p} - \mathbf{H}_p\rVert
                + \lambda_{\rm S} \mathcal{S}(\mathbf{\hat{H}_p}, \mathbf{H}_p)
                + \lambda_{\rm E} \mathcal{E}(\mathbf{\hat{S}_p}, \mathbf{S}_p)
\end{equation}
where $\mathbf{\hat{H}}_p$ and $\mathbf{\hat{S}}_p$ denote the generated height field and semantic map patches, respectively.
$\mathbf{H}_p$ and $\mathbf{S}_p$ are the corresponding ground truth.
The autoregressive transformer in MaskGIT is trained using a reweighted ELBO loss~\cite{DBLP:conf/eccv/Bond-TaylorH0BW22}.

\subsection{City Background Generator}
\label{sec:city-bg-generator}

\noindent \textbf{Scene Representation.}
Similar to SceneDreamer~\cite{DBLP:arxiv/2302-01330}, we use the bird's-eye-view (BEV) scene representation for its efficiency and expressive capabilities, making it easily applicable to unbounded scenes.
Different from 
GANCraft~\cite{DBLP:conf/iccv/HaoMB021} and InfiniCity~\cite{DBLP:conf/iccv/LinLMCSYT23}, 
where features are parameterized to voxel corners, the BEV representation comprises a feature-free 3D volume generated from a height field and a semantic map, following Equation~\ref{eq:layout-gen}.
Specifically, we initiate the process by selecting a local window with a resolution of $N^H_G \times N^W_G \times N^D_G$ from the city layout $\mathbf{L}$.
This local window is denoted as $\mathbf{L}_{\rm G}^{\rm Local}$, which is generated from the corresponding height field $\mathbf{H}_{\rm G}^{\rm Local}$ and semantic map $\mathbf{S}_{\rm G}^{\rm Local}$.

\noindent \textbf{Scene Parameterization.}
To achieve generalizable 3D representation learning across various scenes and align content with 3D semantics, it is necessary to parameterize the scene representation into a latent space, making adversarial learning easier.
For background stuff, we adopt the generative neural hash grid~\cite{DBLP:arxiv/2302-01330} to learn generalizable features across scenes by modeling the hyperspace beyond 3D space.
Specifically, we first encode the local scene $(\mathbf{H}_{\rm G}^{\rm Local}, \mathbf{S}_{\rm G}^{\rm Local})$ using the global encoder $E_{\rm G}$ to produce the compact scene-level feature $\mathbf{f}_{\rm G} \in \mathbb{R}^{d_{\rm G}}$.
\begin{equation}
  \mathbf{f}_{\rm G} = E_{\rm G}(\mathbf{H}_{\rm G}^{\rm Local}, \mathbf{S}_{\rm G}^{\rm Local})
  \label{eq:global-encoder}
\end{equation}
By leveraging a learnable neural hash function $\mathcal{H}$, the indexed feature $\mathbf{f}_{\rm G}^{\mathbf{p}}$ at 3D position $\mathbf{p} \in \mathbb{R}^3$ can be obtained by mapping $\mathbf{p}$ and $\mathbf{f}_{\rm G}$ into a hyperspace, \textit{i.e.}, $\mathbb{R}^{3 + d_G} \rightarrow \mathbb{R}^{N_G^C}$.
\begin{equation}
  \mathbf{f}_{\rm G}^{\mathbf{p}} =
  \mathcal{H}(\mathbf{p}, \mathbf{f}_{\rm G}) = 
      \Big(\bigoplus^{d_G}_{i=1}f_{\rm G}^i\pi^i \bigoplus^3_{j=1}p^j\pi^j \Big) \mod T
  \label{eq:hashgrid}
\end{equation}
where $\oplus$ denotes the bit-wise XOR operation.
$\pi^i$ and $\pi^j$ represent large and unique prime numbers.
We construct $N_H^L$ levels of multi-resolution hash grids to represent multi-scale features, $T$ is the maximum number of entries per level, and $N_G^C$ denotes the number of channels in each unique feature vector.

\noindent \textbf{Volumetric Rendering.}
In a perspective camera model, each pixel in the image corresponds to a camera ray $\mathbf{r}(t) = \mathbf{o}$ + t$\mathbf{v}$, where the ray originates from the center of projection $\mathbf{o}$ and extends in the direction $\mathbf{v}$.
Thus, the corresponding pixel value $C(\mathbf{r})$ is derived from an integral.
\begin{equation}
  C({\mathbf{r}}) =\int^{\infty}_{0}
  T(t)
  \mathbf{c}(\mathbf{f}_{\rm G}^{\mathbf{r}(t)}, l(\mathbf{r}(t)))
  \boldsymbol{\sigma}(\mathbf{f}_{\rm G}^{\mathbf{r}(t)}) dt
\end{equation}
where $T(t) = \mathrm{exp}(-\int^t_0\sigma(\mathbf{f}_{\rm G}^{\mathbf{r}(s)})ds)$.
$l(\mathbf{p})$ represent the semantic label at the 3D position $\mathbf{p}$.
$\mathbf{c}$ and $\boldsymbol{\sigma}$ denote the color and volume density, respectively.

\noindent \textbf{Loss Function.}
The city background generator is trained using a hybrid objective, which includes a combination of a reconstruction loss and an adversarial learning loss.
Specifically, we leverage the L1 loss, perceptual loss $\mathcal{P}$~\cite{DBLP:conf/eccv/JohnsonAF16}, and GAN loss $\mathcal{G}$~\cite{DBLP:arxiv/LimY17} in this combination.
\begin{equation}
  \ell_{\rm G} = \lambda_{\rm L1} \lVert\mathbf{\hat{I}}_{\rm G} - \mathbf{I}_{\rm G}\rVert
               + \lambda_{\rm P} \mathcal{P}(\mathbf{\hat{I}}_{\rm G}, \mathbf{I}_{\rm G})
               + \lambda_{\rm G} \mathcal{G}(\mathbf{\hat{I}}_{\rm G}, \mathbf{S}_{\rm G})
\end{equation}
where $\mathbf{I}_{\rm G}$ denotes the ground truth background image.
$\mathbf{S}_{\rm G}$ is the semantic map in perspective view generated by accumulating semantic labels sampled from the $\mathbf{L}_{\rm G}^{\rm Local}$ along each ray.
The weights of the three losses are denoted as $\lambda_{\rm L1}$, $\lambda_{\rm P}$, and $\lambda_{\rm G}$.
Note that $\ell_{\rm G}$ is solely applied to pixels with semantic labels belonging to background stuff.

\subsection{Building Instance Generator}
\label{sec:building-ins-generator}

\noindent \textbf{Scene Representation.}
Just like the city background generator, the building instance generator also uses the BEV scene representation.
In the building instance generator, we extract a local window denoted as $\mathbf{L}_{\rm B_i}^{\rm Local}$ from the city layout $\mathbf{L}$, with a resolution of $N_B^H \times N_B^W \times N_B^D$, centered around the 2D center $(c_{\rm B_i}^x, c_{\rm B_i}^y)$ of the building instance $B_i$.
The height field and semantic map used to generate $\mathbf{L}_{\rm B_i}^{\rm Local}$ can be denoted as $\mathbf{H}_{\rm B_i}^{\rm Local}$ and $\mathbf{S}_{\rm B_i}^{\rm Local}$, respectively.
As all buildings have the same semantic label in $\mathbf{S}$, we perform building instantiation by detecting connected components.
We observe that the fa\c{c}ades and roofs of buildings in real-world scenes exhibit distinct distributions. 
Consequently, we assign different semantic labels to the fa\c{c}ade and roof of the building instance $\rm B_i$ in $\mathbf{L}_{\rm B_i}^{\rm Local}$, with the top-most voxel layer being assigned the roof label.
The rest building instances are omitted in $\mathbf{L}_{\rm B_i}^{\rm Local}$ by assigned with the null class.

\noindent \textbf{Scene Parameterization.}
In contrast to the city background generator, the building instance generator employs a novel scene parameterization that relies on pixel-level features generated by a local encoder $E_B$.
Specifically, we start by encoding the local scene $(\mathbf{H}_{\rm B_i}^{\rm Local}, \mathbf{S}_{\rm B_i}^{\rm Local})$ using $E_B$, resulting in the pixel-level feature $\mathbf{f}_{\rm B_i}$, which has a resolution of $N_B^H \times N_B^W \times N_B^C$.
\begin{equation}
  \mathbf{f}_{\rm B_i} = E_B(\mathbf{H}_{\rm B_i}^{\rm Local}, \mathbf{S}_{\rm B_i}^{\rm Local})
  \label{eq:local-encoder}
\end{equation}
Given a 3D position $\mathbf{p} = (p_x, p_y, p_z)$, the corresponding indexed feature $\mathbf{f}_{\rm B_i}^{\mathbf{p}}$ can be computed as
\begin{equation}
  \mathbf{f}_{\rm B_i}^{\mathbf{p}} = \mathcal{O}({\rm Concat}(\mathbf{f}_{\rm B_i}^{(p_x, p_y)}, p_z))
\end{equation}
where $\rm Concat(\cdot)$ is the concatenation operation.
$\mathbf{f}_{\rm B_i}^{(p_x, p_y)} \in \mathbb{R}^{N_B^C}$ denotes the feature vector at $(p_x, p_y)$.
$\mathcal{O}(\cdot)$ is the positional encoding function used in the vanilla NeRF~\cite{DBLP:conf/eccv/MildenhallSTBRN20}.
\begin{equation}
  \mathcal{O}(x) = \{\sin(2^i \pi x), \cos(2^i \pi x)\}_{i=0}^{N_P^L - 1}
  \label{eq:sincos}
\end{equation}
Note that $\mathcal{O}(\cdot)$ is applied individually to each value in the given feature $x$, which are normalized to lie within the range of $[-1, 1]$.

\noindent \textbf{Volumetric Rendering.}
Different from the volumetric rendering used in the city background generator, we incorporate a style code $\mathbf{z}$ in the building instance generator to capture the diversity of buildings. 
The corresponding pixel value $C(\mathbf{r})$ is obtained through an integral process.
\begin{equation}
  C({\mathbf{r}}) =\int^{\infty}_{0}
  T(t)
  \mathbf{c}(\mathbf{f}_{\rm B_i}^{\mathbf{r}(t)}, \mathbf{z}, l(\mathbf{r}(t)))
  \boldsymbol{\sigma}(\mathbf{f}_{\rm B_i}^{\mathbf{r}(t)}) dt
\end{equation}
Note that the camera ray $\mathbf{r}(t)$ is normalized with respect to $(c_{\rm B_i}^x, c_{\rm B_i}^y, 0)$ as the origin.

\begin{figure}[!t]
  \centering
  \includegraphics[width=\linewidth]{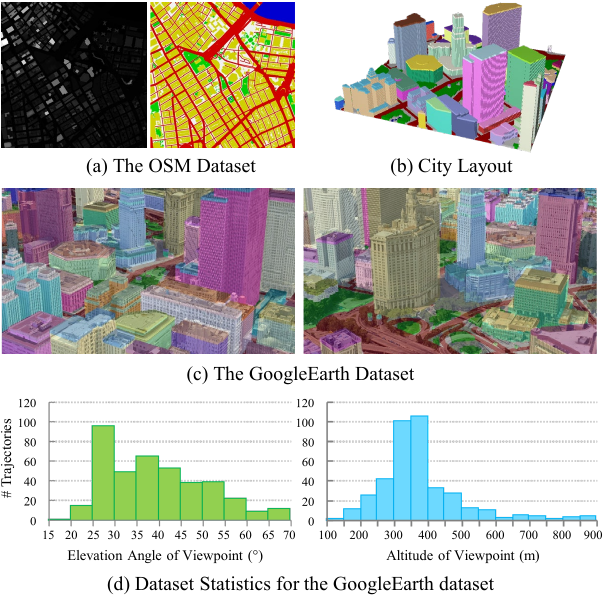}
  \caption{\textbf{Overview of CityGen Datasets.}
  (a) The OSM dataset comprising paired height fields and semantic maps provides real-world city layouts.
  (b) The city layout, generated from the height field and semantic map, facilitates automatic annotation generation.
  (c) The GoogleEarth dataset includes real-world city appearances alongside semantic segmentation and building instance segmentation.
  (d) The dataset statistics demonstrate the variety of perspectives available in the GoogleEarth dataset.}
  \label{fig:dataset-preview}
\end{figure}

\noindent \textbf{Loss Function.}
For training the building instance generator, we exclusively use the GAN Loss. 
Mathematically, it can be represented as
\begin{equation}
  \ell_{\rm B} = \mathcal{G}(\mathbf{\hat{I}}_{\rm B_i}, \mathbf{S}_{\rm B_i})
\end{equation}
where $\mathbf{S}_{\rm B_i}$ denotes the semantic map of building instance $\rm B_i$ in perspective view, which is generated in a similar manner to $\mathbf{S}_{\rm G}$.
Note that $\ell_{\rm B}$ is exclusively applied to pixels with semantic labels belonging to the building instance.

\subsection{Compositor}
\label{sec:compositor}
Since there are no corresponding ground truth images for the images generated by the City Background Generator and Building Instance Generator, it is not possible to train neural networks to merge these images.
Therefore, the compositor uses the generated images $\mathbf{\hat{I}}_{\rm G}$ and $\{\mathbf{\hat{I}}_{\rm B_i}\}_{i=1}^n$, along with their corresponding binary masks $\mathbf{M}_{\rm G}$ and $\{\mathbf{M}_{\rm B_i}\}_{i=1}^n$, the compositor combines them into a unified image $\mathbf{I}_{\rm C}$, which can be represented as
\begin{equation}
  \mathbf{I}_{\rm C} = \mathbf{\hat{I}}_{\rm G}\mathbf{M}_{\rm G}
                     + \sum_{i=1}^n {\mathbf{\hat{I}}_{\rm B_i}\mathbf{M}_{\rm B_i}}
\end{equation}
where $n$ is the number of building instances.

\section{CityGen Datasets}

\noindent \textbf{The OSM Dataset.}
The OSM dataset, sourced from OpenStreetMap~\cite{HAOZHE:link/OpenStreetMap}, is composed of the rasterized semantic maps and height fields of 80 cities worldwide, spanning an area of more than 6,000 km$^2$.
During the rasterization process, vectorized geometry information is converted into images by translating longitude and latitude into the EPSG:3857 coordinate system at zoom level 18, approximately 0.597 meters per pixel.
As shown in Figure~\ref{fig:dataset-preview}\hyperref[fig:dataset-preview]{(a)}, the segmentation maps use red, yellow, green, cyan, and blue colors to denote the positions of roads, buildings, green lands, construction sites, and water areas, respectively.
The height fields primarily represent the height of buildings, with their values derived from OpenStreetMap. 
For roads, the height values are set to 4, while for water areas, they are set to 0.
Additionally, the height values for trees are sampled from perlin noise~\cite{DBLP:conf/siggraph/Perlin85}, ranging from 8 to 16.

\begin{table}[!t]
  \caption{\textbf{A Comparison of GoogleEarth with representative city-related datasets.} Note that the number of images and area are counted based on real-world images. ``sate.'' represents satellite. ``inst.'', ``sem.'', and ``plane'' denote ``instance segmentation'', ``semantic segmentation'', and ``plane segmentation'' respectively.}
  \vspace{-2 mm}
  \resizebox{\linewidth}{!}{
    \begin{tabular}{l|ccccc}
      \toprule
      Dataset  & \# Images & Area           & View         & Annotation  & 3D \\
      \midrule
      KITTI~\cite{DBLP:conf/cvpr/GeigerLU12}
               & 15 k      & -              & street       & sem.        & \xmark \\
      Cityscapes~\cite{DBLP:conf/cvpr/CordtsORREBFRS16}
               & 25 k      & -              & street       & sem.        & \xmark \\
      SpaceNet MOVI~\cite{DBLP:conf/iccv/WeirLBEVMST19}
               & 6.0 k     & -              & sate.        & inst.       & \xmark \\
      OmniCity~\cite{DBLP:conf/cvpr/LiLXXYHXL23}
               & 108 k     & -              & street/sate. & inst./plane & \xmark \\
      \midrule
      HoliCity~\cite{DBLP:arxiv/2008-03286} 
               & 6.3 k     & 20 $\rm km^2$  & street       & sem./plane  & \cmark \\
      UrbanScene3D~\cite{DBLP:conf/eccv/LinLHYXH22}
               & 6.1 k     & 3 $\rm km^2$   & drone        & inst.       & \cmark \\
      GoogleEarth 
               & 24 k      & 25 $\rm km^2$  & drone        & inst./sem.  & \cmark \\
      \bottomrule
    \end{tabular}
  }
  \label{tab:dataset-comparison}
\end{table}

\noindent \textbf{The GoogleEarth Dataset.}
The GoogleEarth dataset is collected from Google Earth Studio~\cite{HAOZHE:link/GoogleEarth}, including 400 orbit trajectories in Manhattan and Brooklyn. 
Each trajectory consists of 60 images, with orbit radiuses ranging from 125 to 813 meters and altitudes varying from 112 to 884 meters.
In addition to the images, Google Earth Studio provides camera intrinsic and extrinsic parameters, making it possible to create automated annotations for semantic and building instance segmentation.
Specifically, for building instance segmentation, we initially perform connected components detection on the semantic maps to identify individual building instances.
Then, the city layout is created following Equation~\ref{eq:layout-gen}, as demonstrated in Figure~\ref{fig:dataset-preview}\hyperref[fig:dataset-preview]{(b)}.
Finally, the annotations are generated by projecting the city layout onto the images, using the camera parameters, as shown in Figure~\ref{fig:dataset-preview}\hyperref[fig:dataset-preview]{(c)}.
Table~\ref{tab:dataset-comparison} presents a comparative overview between GoogleEarth and other datasets related to urban environments.
Among datasets that offer 3D models, GoogleEarth stands out for its extensive coverage of real-world images, encompassing the largest area, and providing annotations for both semantic and instance segmentation.
Figure~\ref{fig:dataset-preview}\hyperref[fig:dataset-preview]{(d)} offers an analysis of viewpoint altitudes and elevations in the GoogleEarth dataset, highlighting its diverse camera viewpoints. 
This diversity enhances neural networks' ability to generate cities from a broader range of perspectives.
Additionally, leveraging Google Earth and OpenStreetMap data allows us to effortlessly expand our dataset to encompass more cities worldwide.

\section{Experiments}

\begin{figure*}[!t]
  \centering
  \includegraphics[width=\linewidth]{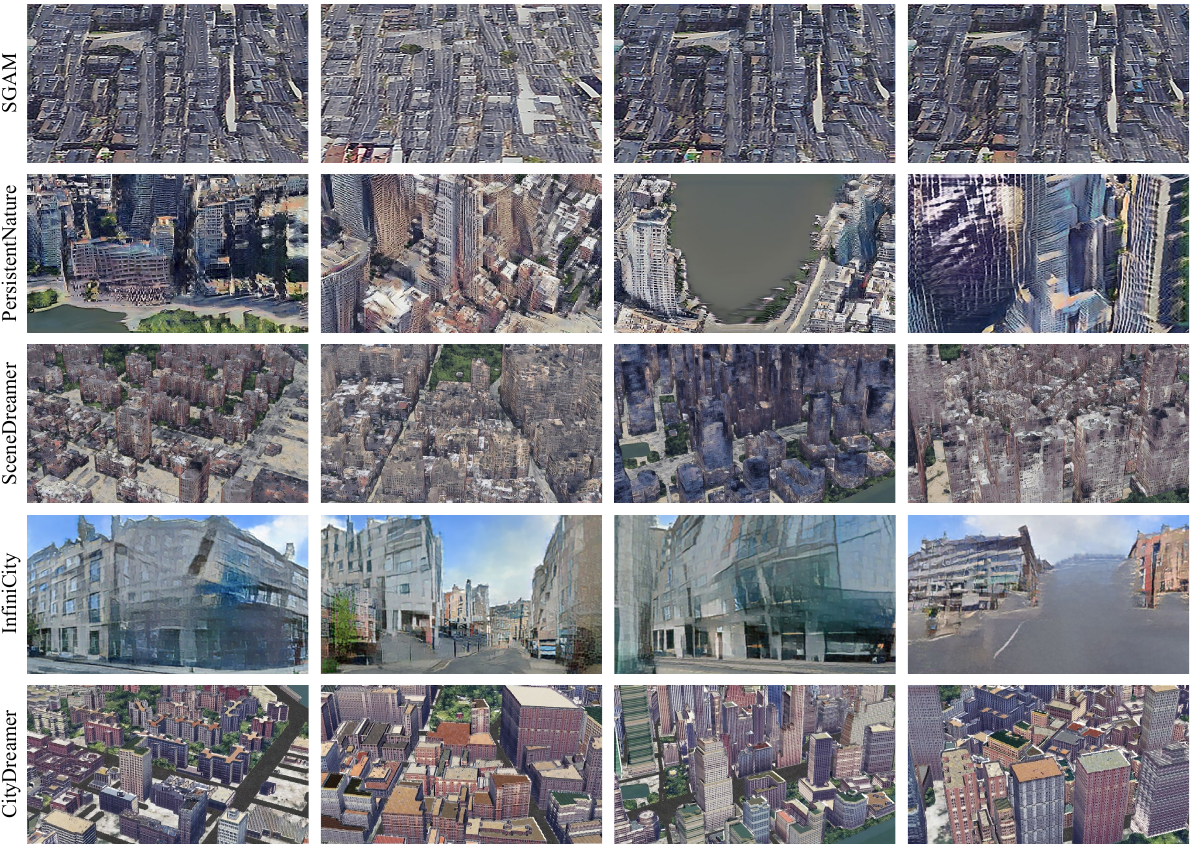}
  \caption{\textbf{Qualitative comparison.} The proposed CityDreamer produces more realistic and diverse results compared to all baselines. Note that the visual results of InfiniCity~\cite{DBLP:conf/iccv/LinLMCSYT23} are provided by the authors and zoomed for optimal viewing.}
  \label{fig:citygen-comparison}
\end{figure*}

\subsection{Evaluation Protocols}

During evaluation, we use the Unbounded Layout Generator to generate 1024 distinct city layouts. 
For each scene, we sample 20 different styles by randomizing the style code $\mathbf{z}$.
Each sample is transformed into a fly-through video consisting of 40 frames, each with a resolution of 960$\times$540 pixels and any possible camera trajectory. Subsequently, we randomly select frames from these video sequences for evaluation.
The evaluation metrics are as follows: 

\noindent \textbf{FID and KID.}
Fr\'echet Inception Distance (FID)~\cite{DBLP:conf/nips/HeuselRUNH17} and Kernel Inception Distance (KID)~\cite{DBLP:conf/iclr/BinkowskiSAG18} are metrics for the quality of generated images.
We compute FID and KID between a set of 15,000 generated frames and an evaluation set comprising 15,000 images randomly sampled from the GoogleEarth dataset.

\noindent \textbf{Depth Error.}
We employ depth error (DE) to assess the 3D geometry, following a similar approach to EG3D~\cite{DBLP:conf/cvpr/ChanLCNPMGGTKKW22}.
Using a pre-trained model~\cite{DBLP:journals/pami/RanftlLHSK22}, we generate pseudo ground truth depth maps for generated frames by accumulating density $\sigma$. 
Both the ``ground truth'' depth and the predicted depth are normalized  to zero mean and unit variance to eliminate scale ambiguity. 
DE is computed as the L2 distance between the two normalized depth maps. We assess this depth error on 100 frames for each evaluated method.

\noindent \textbf{Camera Error.}
Following SceneDreamer~\cite{DBLP:arxiv/2302-01330}, we introduce camera error (CE) to assess multi-view consistency.
CE quantifies the difference between the inference camera trajectory and the estimated camera trajectory from COLMAP~\cite{DBLP:conf/cvpr/SchonbergerF16}.
It is calculated as the scale-invariant normalized L2 distance between the reconstructed and generated camera poses.

\subsection{Implementation Details}


\noindent \textbf{Hyperparameters:}

\noindent \textit{Unbounded Layout Generator.}
The codebook size $K$ is set to 512, and each code's dimension $D$ is set to 512.
The height field and semantic map patches are cropped to a size of 512$\times$512, and compressed by a factor of 16.
The loss weights, $\lambda_{\rm R}$, $\lambda_{\rm S}$, and $\lambda_{\rm E}$, are 10, 10, 1, respectively.

\noindent \textit{City Background Generator.}
The local window resolution $N_G^H$, $N_G^W$, and $N_G^D$ are set to 1536, 1536, and 640, respectively.
The dimension of the scene-level features $d_G$ is 2.
For the generative hash grid, we use $N_H^L = 16$, $T = 2^{19}$, and $N_G^C = 8$. 
The unique prime numbers in Equation~\ref{eq:hashgrid} are set to $\pi^1 = 1$, $\pi^2 = 2654435761$, $\pi^3 = 805459861$, $\pi^4 = 3674653429$, and $\pi^5 = 2097192037$.
The loss function weights, $\lambda_{\rm L1}$, $\lambda_{\rm P}$, and $\lambda_{\rm G}$, are 10, 10, 0.5, respectively.

\noindent \textit{Building Instance Generator.}
The local window resolution $N_B^H$, $N_B^W$, and $N_B^D$ are set to 672, 672, and 640, respectively.
The number of channels $N_B^C$ of the pixel-level features is 63.
The dimension $N_P^L$ is set to 10.

\noindent \textbf{Training Details:}

\noindent \textit{Unbounded Layout Generator.}
The VQVAE is trained with a batch size of 16 using an Adam optimizer with $\beta$ = (0.5, 0.9) and a learning rate of $7.2\times10^{-5}$ for 1,250,000 iterations.
The autoregressive transformer is trained with a batch size of 80 using an Adam optimizer with $\beta$ = (0.9, 0.999) and a learning rate of $2\times10^{-4}$ for 250,000 iterations.

\noindent \textit{City Background and Building Instance Generators.}
Both generators are trained using an Adam optimizer with $\beta$ = (0, 0.999) and a learning rate of $10^{-4}$.
The discriminators are optimized using an Adam optimizer with $\beta$ = (0, 0.999) and a learning rate of $10^{-5}$.
The training lasts for 298,500 iterations with a batch size of 8.
The images are randomly cropped to a size of 192$\times$192.

\begin{table}[!t]
  \centering
  \caption{\textbf{Quantitative comparison}. The best values are highlighted in bold. Note that the results of InfiniCity are not included in this comparison as it is not open-sourced.}
  \vspace{-2 mm}
  \resizebox{\linewidth}{!}{
    \begin{tabular}{l|c|c|c|c}
      \toprule
      Methods & FID $\downarrow$
              & KID $\downarrow$
              & DE $\downarrow$
              & CE $\downarrow$ \\
      \midrule
      SGAM~\cite{DBLP:conf/nips/ShenMW22} 
              & 277.64     & 0.358      & 0.575      & 239.291 \\
      PersistentNature~\cite{DBLP:conf/cvpr/ChaiTLIS23} 
              & 123.83     & 0.109      & 0.326      & 86.371 \\
      SceneDreamer~\cite{DBLP:arxiv/2302-01330} 
              & 213.56     & 0.216      & 0.152      & 0.186 \\
      CityDreamer 
              & \bf{97.38} & \bf{0.096} & \bf{0.147} & \bf{0.060} \\
      \bottomrule
    \end{tabular}
  }
  \label{tab:quantative-cmp}
\end{table}

\subsection{Main Results}

\noindent \textbf{Comparison Methods.}
We compare CityDreamer against four state-of-the-art methods: SGAM~\cite{DBLP:conf/nips/ShenMW22}, PersistentNature~\cite{DBLP:conf/cvpr/ChaiTLIS23}, SceneDreamer~\cite{DBLP:arxiv/2302-01330}, and InfiniCity~\cite{DBLP:conf/iccv/LinLMCSYT23}.
With the exception of InfiniCity, whose code is not available, the remaining methods are retrained using the released code on the GoogleEarth dataset to ensure a fair comparison.
SceneDreamer initially uses simplex noise for layout generation, which is not ideal for cities, so it is replaced with the unbounded layout generator from CityDreamer.

\noindent \textbf{Qualitative Comparison.}
Figure~\ref{fig:citygen-comparison} provides qualitative comparisons against baselines.
SGAM struggles to produce realistic results and maintain good 3D consistency because extrapolating views for complex 3D cities can be extremely challenging.
PersistentNature employs tri-plane representation, but it encounters challenges in generating realistic renderings.
SceneDreamer and InfiniCity both utilize voxel grids as their representation, but they still suffer from severe structural distortions in buildings because all buildings are given the same semantic label.
In comparison, the proposed CityDreamer generates more realistic and diverse results compared to all the baselines.

\noindent \textbf{Quantitative Comparison.}
Table~\ref{tab:quantative-cmp} presents the quantitative metrics of the proposed approach compared to the baselines.
CityDreamer exhibits significant improvements in FID and KID, which is consistent with the visual comparisons.
Moreover, CityDreamer demonstrates the capability to maintain accurate 3D geometry and view consistency while generating photorealistic images, as evident by the lowest DE and CE errors compared to the baselines.

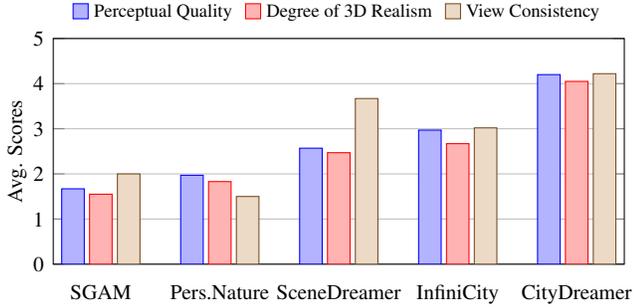
\begin{figure}[!t]
  \begin{tikzpicture}
  \pgfplotstableread[row sep=\\,col sep=&]{
    Method        & PQ   & DR   & VC \\
    SGAM          & 1.67 & 1.55 & 2.00  \\
    Pers.Nature   & 1.97 & 1.83 & 1.50  \\
    SceneDreamer  & 2.57 & 2.47 & 3.67  \\
    InfiniCity    & 2.97 & 2.67 & 3.02  \\
    CityDreamer   & 4.20 & 4.05 & 4.22  \\
  }\userStudyData

  \begin{axis}[
    ybar,
    width             = \linewidth,
    height            = .55\linewidth,
    ymajorgrids       = true,
    symbolic x coords = {
      SGAM,
      Pers.Nature,
      SceneDreamer,
      InfiniCity,
      CityDreamer
    },
    ymin              = 0,
    ymax              = 5,
    x                 = 10 ex,
    bar width         = 3 mm,
    ylabel            = {Avg. Scores},
    ylabel style      = {
      font            = \fontsize{8}{8}\selectfont
    },
    y label style     = {at = {(-0.03, 0.5)}},
    x tick style      = {draw = none},
    xticklabel style  = {
      align           = center,
      font            = \fontsize{8}{8}\selectfont,
    },
    ytick             = {0, 1, 2, 3, 4, 5},
    yticklabels       = {0, 1, 2, 3, 4, 5},
    yticklabel style  = {
      align           = center,
      font            = \fontsize{8}{8}\selectfont,
    },
    legend image code/.code={
      \draw [#1] (0 mm, -0.8 mm) rectangle (2 mm, 1.2 mm);
    },
    legend style      = {
      at              = {(0.5, 1.2)},
      anchor          = north,
      draw            = none,
      font            = \fontsize{7}{7}\selectfont,
      legend columns  = -1,
      /tikz/every even column/.append style={column sep = 1 mm}
    }]
    \addplot table[x = Method,y = PQ]{\userStudyData};
    \addplot table[x = Method,y = DR]{\userStudyData};
    \addplot table[x = Method,y = VC]{\userStudyData};

    \addlegendentry{Perceptual Quality};
    \addlegendentry{Degree of 3D Realism};
    \addlegendentry{View Consistency};
  \end{axis}
\end{tikzpicture}
  \vspace{-5 mm}
  \caption{\textbf{User study on unbounded 3D city generation.} All scores are in the range of 5, with 5 indicating the best. Note that ``Pers.Nature'' denotes ``PersistentNature''~\cite{DBLP:conf/cvpr/ChaiTLIS23}.}
  \label{fig:user-study}
  \vspace{-3 mm}
\end{figure}

\noindent \textbf{User Study.}
To better assess the 3D consistency and quality of the unbounded 3D city generation, we conduct an output evaluation~\cite{DBLP:journals/ftcgv/BylinskiiHHHZ23} as the user study. 
In this survey, we ask 20 volunteers to rate each generated camera trajectory based on three aspects: 
1) the perceptual quality of the imagery, 2) the level of 3D realism, and 3) the 3D view consistency. 
The scores are on a scale of 1 to 5, with 5 representing the best rating.
The results are presented in Figure~\ref{fig:user-study}, showing that the proposed method significantly outperforms the baselines by a large margin.

\subsection{Ablation Study}

\noindent \textbf{Effectiveness of Unbounded Layout Generator.}
The Unbounded Layout Generator plays a critical role in generating ``unbounded'' city layouts.
We compare it with InfinityGAN~\cite{DBLP:conf/iclr/LinLCT022} used in InfiniCity and a rule-based city layout generation method, IPSM~\cite{DBLP:journals/tog/ChenEWMZ08}, as shown in Table~\ref{tab:ablation-building-generator}.
Following InfiniCity~\cite{DBLP:conf/iccv/LinLMCSYT23}, we use FID and KID to evaluate the quality of the generated layouts.
Compared to IPSM and InfinityGAN, Unbounded Layout Generator achieves better results in terms of all metrics.
The qualitative results shown in Figure~\ref{fig:ablation-layout-generator} in the supplementary material also demonstrate the effectiveness of the proposed method.

\noindent \textbf{Effectiveness of Building Instance Generator.}
We emphasize the crucial role of the building instance generator in the success of unbounded 3D city generation. To demonstrate its effectiveness, we conducted an ablation study on the building instance generator.
We compared two optional designs: (1) Removing the building instance generator from CityDreamer, \textit{i.e.}, the model falling back to SceneDreamer. (2) All buildings are generated at once by the building instance generator, without providing any instance labels.
The quantitative results presented in Table~\ref{tab:ablation-building-generator} demonstrate the effectiveness of both the instance labels and the building instance generator.
Refer to Figure~\ref{fig:ablation-building-generator} in the supplementary material for more qualitative comparisons.

\noindent \textbf{Effectiveness of Scene Parameterization.}
Scene parameterization directly impacts the quality of 3D city generation. 
The city background generator utilizes HashGrid with patch-wise features from the global encoder, while the building instance generator uses vanilla SinCos positional encoding with pixel-wise features from the local encoder. 
We compare different scene parameterizations in both the city background generator and the building instance generator.
Table~\ref{tab:ablation-scene-parameterization} shows that using local encoders in background generation or using global encoders in building generation leads to considerable degradation in image quality, indicated by poor metrics. 
According to Equation~\ref{eq:hashgrid}, the output of HashGrid is determined by the scene-level features and 3D position.
While HashGrid enhances the multi-view consistency of the generated background, it also introduces challenges in building generation, leading to less structurally reasonable buildings.
In contrast, the inherent periodicity of SinCos makes it easier for the network to learn the periodicity of building fa\c{c}ades, leading to improved results in building generation.
Refer to Sec.~\ref{sec:scene-parameterization-discussion} in the supplementary material for a detailed discussion.

\subsection{Further Discussions}

\noindent \textbf{Applications.}
This research primarily benefits applications that require efficient content creation, with notable examples being the entertainment industry.
There is a strong demand to generate content for computer games and movies within this field.

\noindent \textbf{Limitations.}
\textbf{1)} The generation of the city layout involves raising voxels to a specific height, which means that concave geometries like caves and tunnels cannot be modeled and generated.
\textbf{2)} During the inference process, the buildings are generated individually, resulting in a slightly higher computation cost. Exploring ways to reduce the inference cost would be beneficial for future work.

\begin{table}[!t]
  \centering
  \caption{\textbf{Effectiveness of Ubounded Layout Generator.} The best values are highlighted in bold. 
  The images are centrally cropped to a size of 4096$\times$4096.}
  \vspace{-2 mm}
  \begin{tabularx}{\linewidth}{l|Y|Y}
     \toprule
     Methods     & FID $\downarrow$        
                 & KID $\downarrow$ \\ 
     \midrule
     IPSM~\cite{DBLP:journals/tog/ChenEWMZ08}
                 & 321.47      & 0.502      \\ 
     InfinityGAN~\cite{DBLP:conf/iclr/LinLCT022}
                 & 183.14      & 0.288      \\ 
     Ours        & \bf{124.45} & \bf{0.123} \\ 
     \bottomrule
  \end{tabularx}
  \label{tab:ablation-layout-generator}
\end{table}

\begin{table}[!t]
  \centering
  \caption{\textbf{Effectiveness of Building Instance Generator.} The best values are highlighted in bold. Note that ``w/o BIG.'' indicates the removal of Building Instance Generator from CityDreamer. ``w/o Ins.'' denotes the absence of building instance labels in the building instance generator.}
  \vspace{-2 mm}
  \begin{tabularx}{\linewidth}{l|Y|Y|Y|Y}
     \toprule
     Methods  & FID $\downarrow$
              & KID $\downarrow$
              & DE $\downarrow$
              & CE $\downarrow$ \\
     \midrule
     w/o BIG. & 213.56     & 0.216      & 0.152      & 0.186 \\
     w/o Ins. & 117.75     & 0.124      & 0.148      & 0.098 \\
     Ours     & \bf{97.38} & \bf{0.096} & \bf{0.147} & \bf{0.060} \\
     \bottomrule
  \end{tabularx}
  \label{tab:ablation-building-generator}
\end{table}

\begin{table}[!t]
  \centering
  \caption{\textbf{Effectiveness of different generative scene parameterization.} The best values are highlighted in bold. Note that ``CBG.'' and ``BIG.'' denote City Background Generator and Building Instance Generator, respectively. ``Enc.'' and ``P.E.'' represent ``Encoder'' and ``Positional Encoding'', respectively.}
  \vspace{-2 mm}
  \resizebox{\linewidth}{!}{
    \begin{tabular}{c|c|c|c|c|c|c|c}
      \toprule
      \multicolumn{2}{c|}{CBG.} & 
      \multicolumn{2}{c|}{BIG.} &
      \multirow{2}{*}{FID $\downarrow$} & 
      \multirow{2}{*}{KID $\downarrow$} & 
      \multirow{2}{*}{DE $\downarrow$}  & 
      \multirow{2}{*}{CE $\downarrow$} \\
      \cline{1-4}
      Enc.       & P.E.       & Enc.       & P.E.       
                 &            &            &            \\
      \midrule
      Local      & SinCos     & Global     & Hash       &
      219.30     & 0.233      & 0.154      & 0.452 \\
      Local      & SinCos     & Local      & SinCos     &
      107.63     & 0.125      & 0.149      & 0.078 \\
      Global     & Hash       & Global     & Hash       &
      213.56     & 0.216      & 0.153      & 0.186 \\
      Global     & Hash       & Local      & SinCos     &
      \bf{97.38} & \bf{0.096} & \bf{0.147} & \bf{0.060} \\
      \bottomrule
    \end{tabular}
  }
  \label{tab:ablation-scene-parameterization}
  \vspace{-4 mm}
\end{table}

\section{Conclusion}

In this paper, we propose CityDreamer, a compositional generative model designed specifically for unbounded 3D cities.
Compared to existing methods that treat buildings as a single class of objects, CityDreamer separates the generation of building instances from background stuff, allowing for better handling of the diverse appearances of buildings.
Additionally, we create a suite of CityGen Datasets, including OSM and GoogleEarth, providing more realistic city layouts and appearances, and easily scalable to include other cities worldwide.
CityDreamer achieves state-of-the-art performance not only in generating realistic 3D cities but also in localized editing within the generated cities.

\noindent \textbf{Acknowledgments}
%
This study is supported by the Ministry of Education, Singapore, under its MOE AcRF Tier 2 (MOE-T2EP20221-0012), NTU NAP, and under the RIE2020 Industry Alignment Fund – Industry Collaboration Projects (IAF-ICP) Funding Initiative, as well as cash and in-kind contribution from the industry partner(s).

{\small
\bibliographystyle{ieee_fullname}
\bibliography{references}
}

\clearpage
\onecolumn
\appendix

In this supplementary material, we offer extra details and additional results to complement the main paper.
Firstly, we offer more extensive information and results regarding the ablation studies in Sec.~\ref{sec:ablation-study-details}.
Secondly, we present additional experimental results in Sec.~\ref{sec:more-experimental-results}.
Finally, we provide a brief overview of our interactive demo in Sec.~\ref{sec:interactive-demo}.

\section{Additional Ablation Study Results}
\label{sec:ablation-study-details}

\subsection{Qulitative Results for Ablation Studies}

\begin{figure}[!h]
  \centering
  \resizebox{\linewidth}{!}{
    \includegraphics{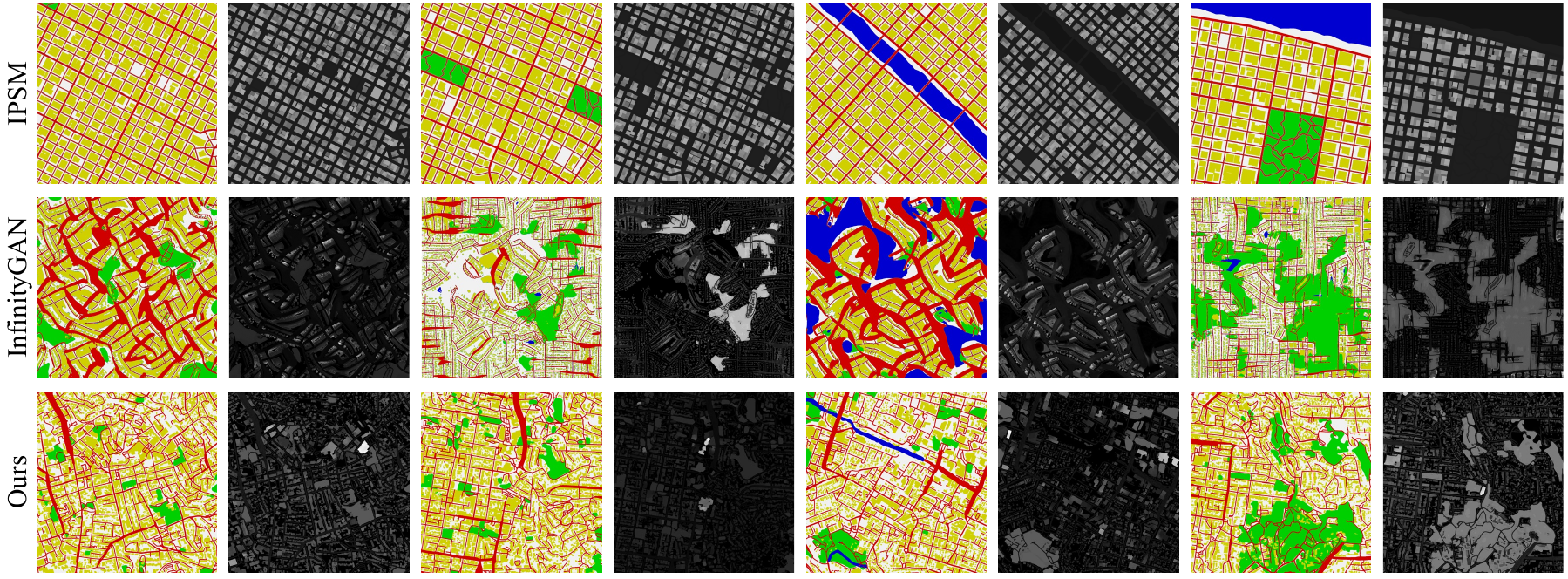}
  }
  \caption{\textbf{Qualitative comparison of different city layout generation methods.} The height map values are normalized to a range of $[0, 1]$ by dividing each value by the maximum value within the map.}
  \label{fig:ablation-layout-generator}
\end{figure}

\noindent \textbf{Effectiveness of Unbounded Layout Generator.}
Figure~\ref{fig:ablation-layout-generator} gives a qualitative comparison as a supplement to Table~\ref{tab:ablation-layout-generator}, demonstrating the effectiveness of Unbounded Layout Generator.
In the case of InfinityGAN, we follow the approach used in InfiniCity, where each class of semantic maps is assigned a specific color, and we convert back to a semantic map by associating it with the nearest color.

\begin{figure}[!h]
  \centering
  \resizebox{\linewidth}{!}{
    \includegraphics{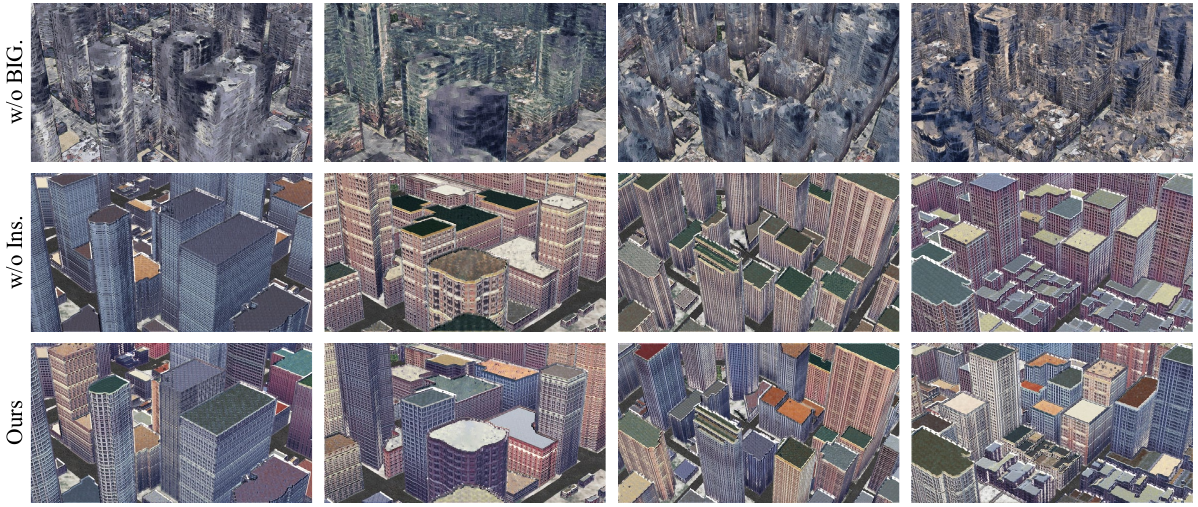}
  }
  \caption{\textbf{Qualitative comparison of different Building Instance Generator variants.} Note that "w/o BIG." indicates the removal of Building Instance Generator from CityDreamer. "w/o Ins." denotes the absence of building instance labels in Building Instance Generator.}
  \label{fig:ablation-building-generator}
  \vspace{-8 mm}
\end{figure}

\noindent \textbf{Effectiveness of Building Instance Generator.}
Figure~\ref{fig:ablation-building-generator} provides a qualitative comparison as a supplement to Table~\ref{tab:ablation-building-generator}, demonstrating the effectiveness of Building Instance Generator.
Figure~\ref{fig:ablation-building-generator} highlights the importance of both Building Instance Generator and the instance labels.
Removing either of them significantly degrades the quality of the generated images.

\clearpage
\subsection{More Discussions on Scene Parameterization}
\label{sec:scene-parameterization-discussion}

Table~\ref{tab:ablation-scene-parameterization} displays the four primary combinations of different encoders and positional encodings.
Additionally, Table~\ref{tab:ablation-scene-parameterization-cont} presents twelve additional alternative combinations, in addition to those in Table~\ref{tab:ablation-scene-parameterization}.
The results in Table~\ref{tab:ablation-scene-parameterization-cont} clearly demonstrate the superiority of the scene parameterization used in CityDreamer.

\begin{figure}[!h]
  \centering
  \resizebox{\linewidth}{!}{
    \includegraphics{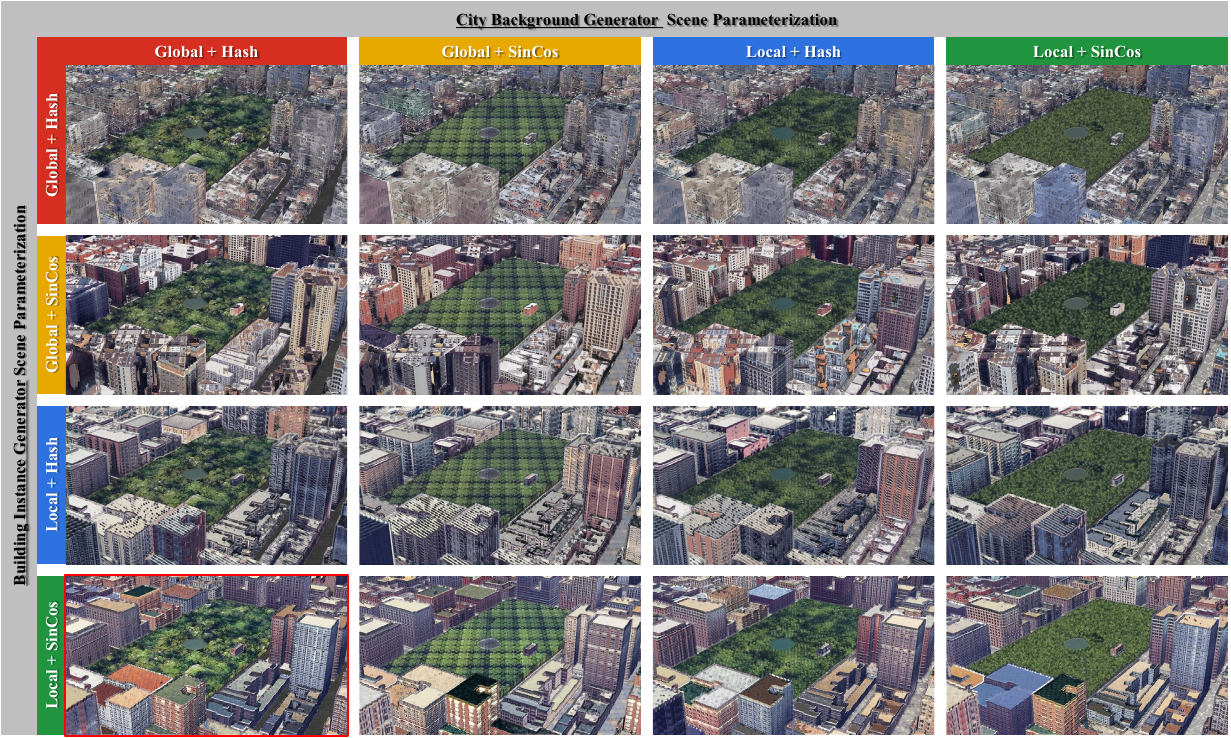}
  }
  \vspace{-6 mm}
  \caption{\textbf{Qualitative comparison of different scene parameterization.} The terms ``Global'' and ``Local'' correspond to ``Global Encoder'' ($E_G$) and "Local Encoder" ($E_B$), which generate features following Equation~\ref{eq:global-encoder} and Equation~\ref{eq:local-encoder} respectively. ``Hash'' and ``SinCos'' represent ``Hash Grid'' and ``SinCos'' positional encodings defined in Equations~\ref{eq:hashgrid} and~\ref{eq:sincos}, respectively.}
  \label{fig:ablation-scene-parameterization}
  \vspace{-6 mm}
\end{figure}

We present the qualitative results for the sixteen scene parameterization settings in Figure~\ref{fig:ablation-scene-parameterization}. 
Using the Global Encoder and Hash Grid as scene parameterization results in more natural city backgrounds (first column) but leads to a severe decrease in the quality of generated buildings (first row). 
As demonstrated in the third row and third column, this irregularity is weakened when the Global Encoder is replaced with the Local Encoder.
Furthermore, using the Global Encoder with SinCos positional encoding introduces periodic patterns, as shown in the second row and second column. 
However, this periodicity is disrupted when the Global Encoder is replaced with the Local Encoder (the fourth row and column) because the input of SinCos positional encoding no longer depends on 3D position $\mathbf{p}$. 
Nevertheless, this change also slightly reduces the multi-view consistency.

\begin{table}[!h]
  \caption{\textbf{Effectiveness of different generative scene parameterization.} The best values are highlighted in bold. Note that ``CBG.'' and ``BIG.'' denote City Background Generator and Building Instance Generator, respectively. ``Enc.'' and ``P.E.'' represent ``Encoder'' and ``Positional Encoding'', respectively.}
  \vspace{-2 mm}
  \centering
  \resizebox{\linewidth}{!}{
    \begin{tabular}{c|c|c|c|c|c|c|c|c|c|c|c|c|c|c|c|c|c}
      \toprule
      \parbox[t]{2mm}{\multirow{2}{*}{\rotatebox[origin=c]{90}{CBG.}}}
      & Enc. & \multicolumn{8}{c|}{Global}        & \multicolumn{8}{c}{Local} \\
      \cline{2-18}
      & P.E. & \multicolumn{4}{c|}{Hash}          & \multicolumn{4}{c|}{SinCos}
             & \multicolumn{4}{c|}{Hash}          & \multicolumn{4}{c}{SinCos} \\
      \midrule
      \parbox[t]{2mm}{\multirow{2}{*}{\rotatebox[origin=c]{90}{BIG.}}}
      & Enc. & \multicolumn{2}{c|}{Global}        & \multicolumn{2}{c|}{Local}
             & \multicolumn{2}{c|}{Global}        & \multicolumn{2}{c|}{Local}
             & \multicolumn{2}{c|}{Global}        & \multicolumn{2}{c|}{Local}
             & \multicolumn{2}{c|}{Global}        & \multicolumn{2}{c}{Local} \\
      \cline{2-18}
      & P.E. & Hash  & SinCos  & Hash   & SinCos     & Hash       & SinCos & Hash   & SinCos
             & Hash  & SinCos  & Hash   & SinCos     & Hash       & SinCos & Hash   & SinCos \\
      \midrule
      \multicolumn{2}{c|}{FID $\downarrow$} &  
              213.56 & 113.45  & 112.61 & \bf{97.38} & 248.30     & 135.86 & 125.97 & 132.67 & 
              203.97 & 116.01  & 116.76 & 99.78      & 219.30     & 124.87 & 137.99 & 107.63 \\
      \multicolumn{2}{c|}{KID $\downarrow$} &
              0.216  & 0.141   & 0.129  & \bf{0.096} & 0.318      & 0.205  & 0.172  & 0.174 & 
              0.199  & 0.105   & 0.104  & 0.098      & 0.233      & 0.134  & 0.157  & 0.125 \\
      \multicolumn{2}{c|}{DE $\downarrow$}  &
              0.153  & 0.149   & 0.153  & \bf{0.147} & 0.156      & 0.155  & 0.150  & 0.151 & 
              0.156  & 0.150   & 0.152  & 0.152      & 0.154      & 0.152  & 0.153  & 0.149 \\
      \multicolumn{2}{c|}{CE $\downarrow$}  &
              0.186  & 0.086   & 0.095  & \bf{0.060} & 0.325      & 0.106  & 0.165  & 0.089 & 
              0.153  & 0.933   & 0.127  & 0.075      & 0.452      & 0.174  & 0.246  & 0.078 \\
      \bottomrule
    \end{tabular}
  }
  \label{tab:ablation-scene-parameterization-cont}
  \vspace{-4 mm}
\end{table}

\clearpage
\section{Additional Experimental Results}
\label{sec:more-experimental-results}

\subsection{View Consistency Comparison}

\begin{figure}[!h]
  \centering
  \resizebox{\linewidth}{!}{
    \includegraphics{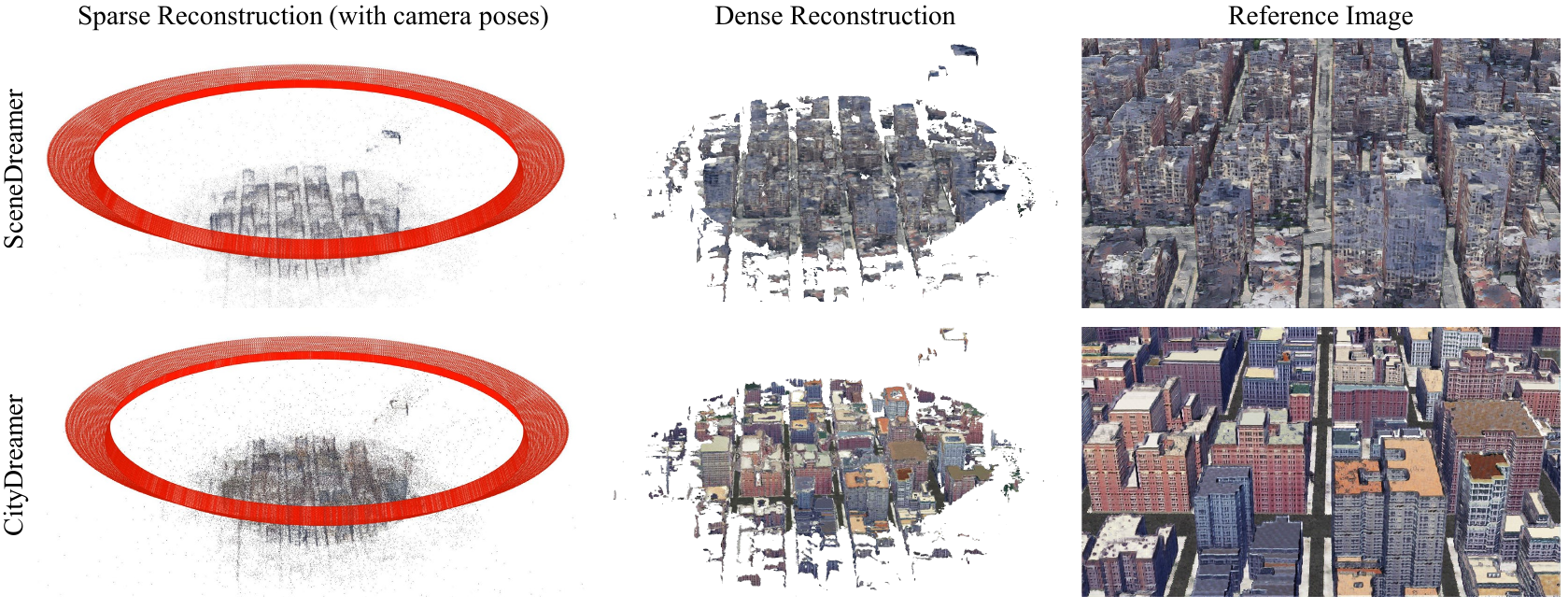}
  }  
  \caption{\textbf{COLMAP reconstruction of a 600-frame generated video captured from an orbit trajectory.} The red ring represents the estimated camera poses, and the well-defined point clouds showcase CityDreamer's highly multi-view consistent renderings.}
  \label{fig:colmap-reconstruction}
\end{figure}

To demonstrate the multi-view consistent renderings of CityDreamer, we utilize COLMAP~\cite{DBLP:conf/cvpr/SchonbergerF16} for structure-from-motion and dense reconstruction using a generated video sequence. 
The video sequence consists of 600 frames with a resolution of 960$\times$540, captured from a circular camera trajectory that orbits around the scene at a fixed height and looks at the center (similar to the sequence presented in the supplementary video). 
The reconstruction is performed solely using the images, without explicitly specifying camera parameters.
As shown in Figure~\ref{fig:colmap-reconstruction}, the estimated camera poses precisely match our sampled trajectory, and the resulting point cloud is well-defined and dense. 
Out of the evaluated methods, only SceneDreamer and CityDreamer managed to accomplish dense reconstruction.
CityDreamer, in particular, exhibited superior view consistency compared to SceneDreamer.
This superiority can be attributed to the fact that the images generated by CityDreamer are more conducive to feature matching.

\subsection{Building Interpolation}

\begin{figure}[!h]
  \centering
  \resizebox{\linewidth}{!}{
    \includegraphics{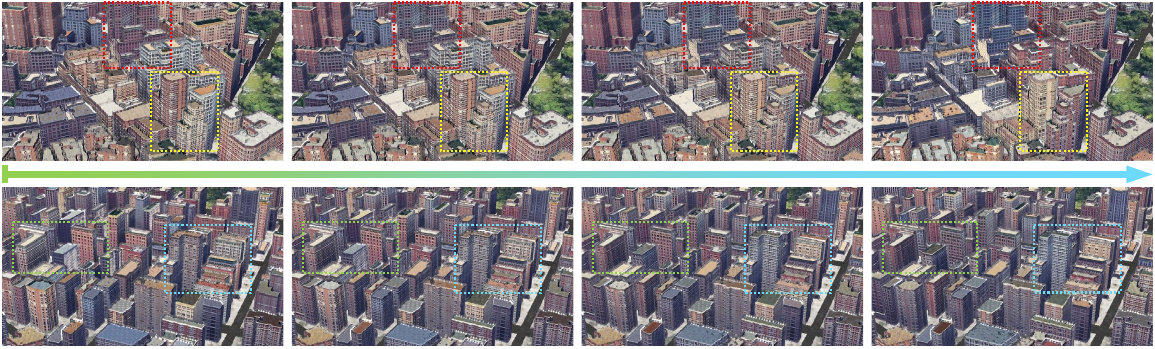}
  }  
  \caption{\textbf{Linear interpolation along the building style.} As we move from left to right, the style of each building changes gradually, while the background remains unchanged.}
  \label{fig:building-interpolation}
\end{figure}

As illustrated in Figure~\ref{fig:building-interpolation}, CityDreamer demonstrates the ability to interpolate along the building style, which is controlled by the variable $\mathbf{z}$.

\clearpage
\subsection{Localized Editing}

\begin{figure}[!h]
  \centering
  \resizebox{\linewidth}{!}{
    \includegraphics{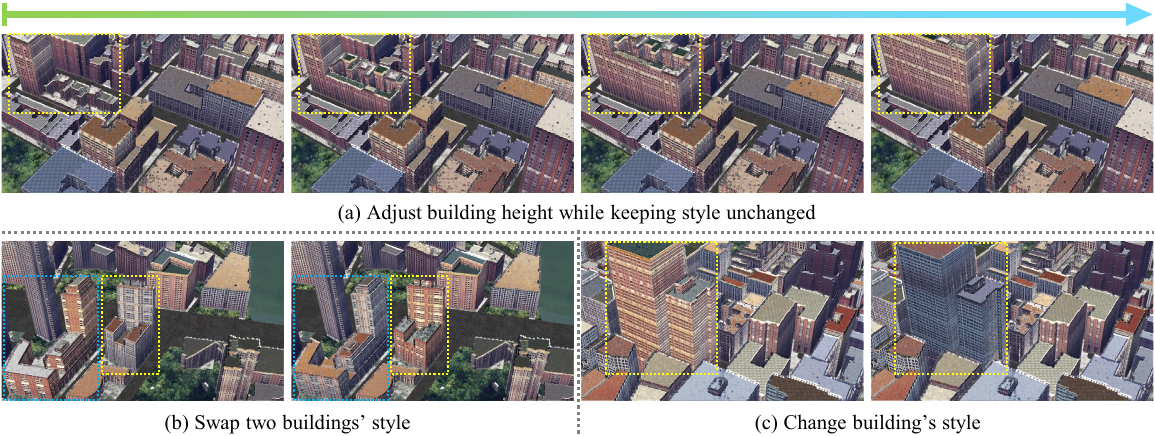}
  }  
  \caption{\textbf{Localized editing for the building instances highlighted with bounding boxes.} (a) While transitioning from left to right, the building's style remains constant, yet its appearance dynamically adjusts to varying heights. (b) The styles of the two buildings can be interchanged. (c) A new style vector can be applied to alter the building's appearance.}
  \label{fig:building-local-edit}
\end{figure}

Benefiting from the compositional architecture, CityDreamer allows for localized editing on building instances.
As shown in Figure~\ref{fig:building-local-edit}, the style and height of each building instance can be independently modified.

\subsection{Relighting}

\begin{figure}[!h]
  \centering
  \resizebox{\linewidth}{!}{
    \includegraphics{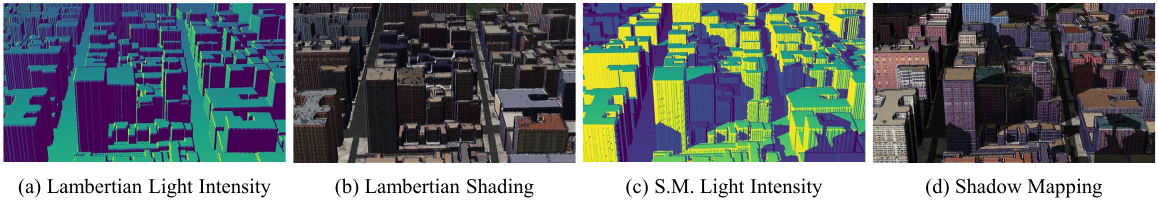}
  }  
  \caption{\textbf{Relighting effects with Lambertian shading and shadow mapping.} (a) and (c) are the light intensity maps. (b) and (d) are the relighted images. Note that ``S.M.'' denotes ``Shadow Mapping''.}
  \label{fig:relighting}
\end{figure}

In CityDreamer, the generation of background stuff and buildings is deliberately decoupled, bringing two advantages: 
\textbf{(1)} Facilitating easier learning for buildings and backgrounds.
\textbf{(2)} Allowing perform local editing on building instances.
The process can be regarded as an inverse rendering, where CityDreamer generates the albedo, normals, and depth of city scenes.
The lighting and shading effects can be subsequently computed based on the provided lighting conditions.
Figure~\ref{fig:relighting} shows the shading effects with Lambertian shading and shadow mapping.
Lambertian shading accounts for the light direction and surface normal, resulting in uniform lighting across all directions, as illustrated in Figures~\ref{fig:relighting}\textcolor{red}{(a)} and \textcolor{red}{(b)}.
The camera is positioned on the left side of the scene.
Shadow mapping further considers light visibility, allowing for the simulation of shadows and occlusion caused by other objects in the scene.
This is shown in Figures~\ref{fig:relighting}\textcolor{red}{(c)} and \textcolor{red}{(d)}. 
The camera is placed at the left rear of the scene.

\clearpage
\subsection{Additional Dataset Examples}

\begin{figure}[!h]
  \centering
  \resizebox{\linewidth}{!}{
    \includegraphics{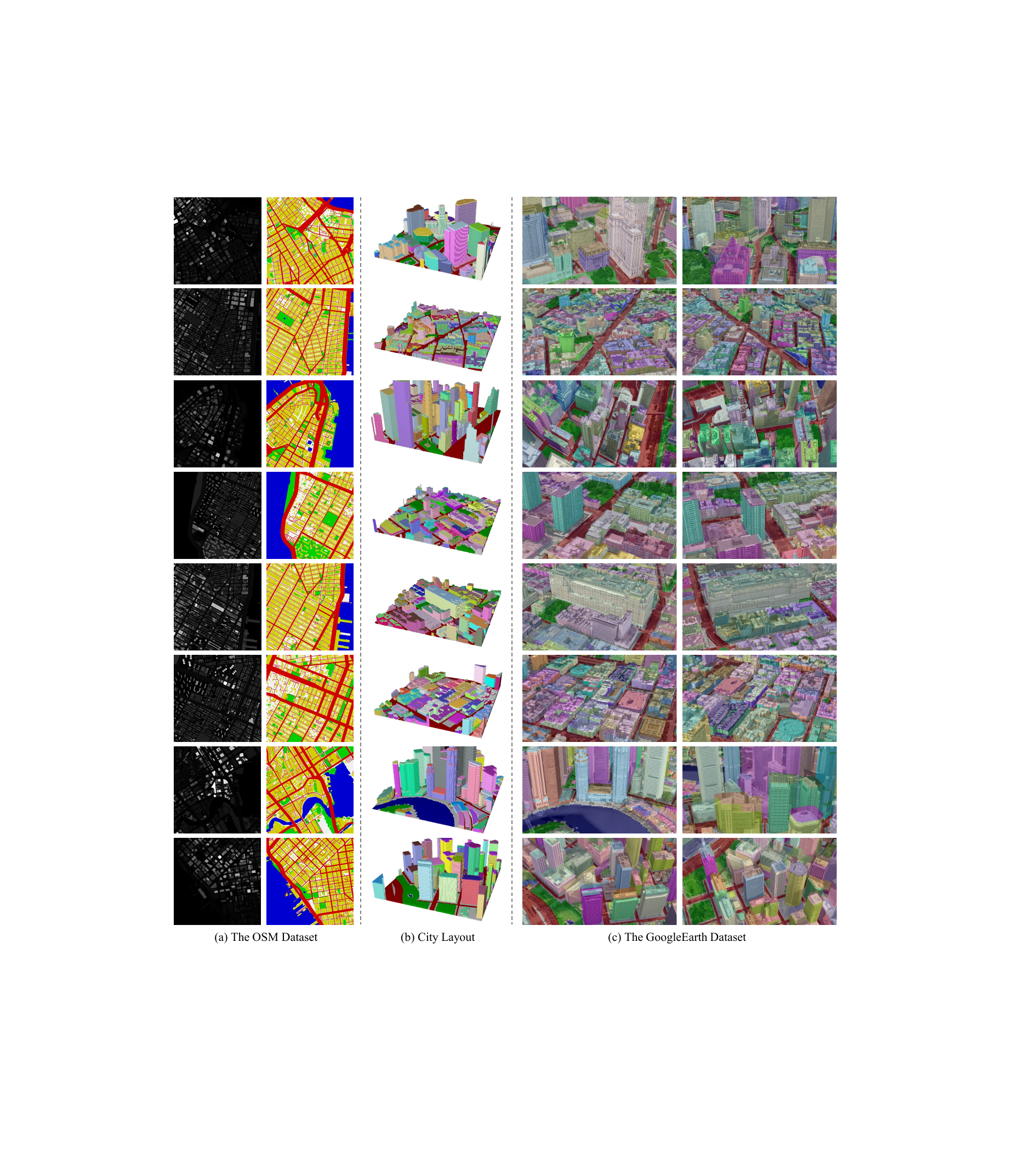}
  }
  \vspace{-6 mm}
  \caption{\textbf{Examples from the OSM and GoogleEarth datasets.}
  (a) Height fields and semantic maps from the OSM dataset.
  (b) City layouts generated from the height fields and semantic maps.
  (c) Images and segmentation maps from the GoogleEarth dataset.}
  \label{fig:dataset-examples-more}
  \vspace{-6 mm}
\end{figure}

In Figure~\ref{fig:dataset-examples-more}, we provide more examples of the OSM and GoogleEarth datasets.
The first six rows are taken from the GoogleEarth dataset, specifically from New York City. 
The last two rows showcase Singapore and San Francisco, illustrating the potential to extend the existing data to other cities worldwide.

\clearpage
\subsection{Additional Qualitative Comparison}

\begin{figure}[!h]
  \centering
  \resizebox{\linewidth}{!}{
    \includegraphics{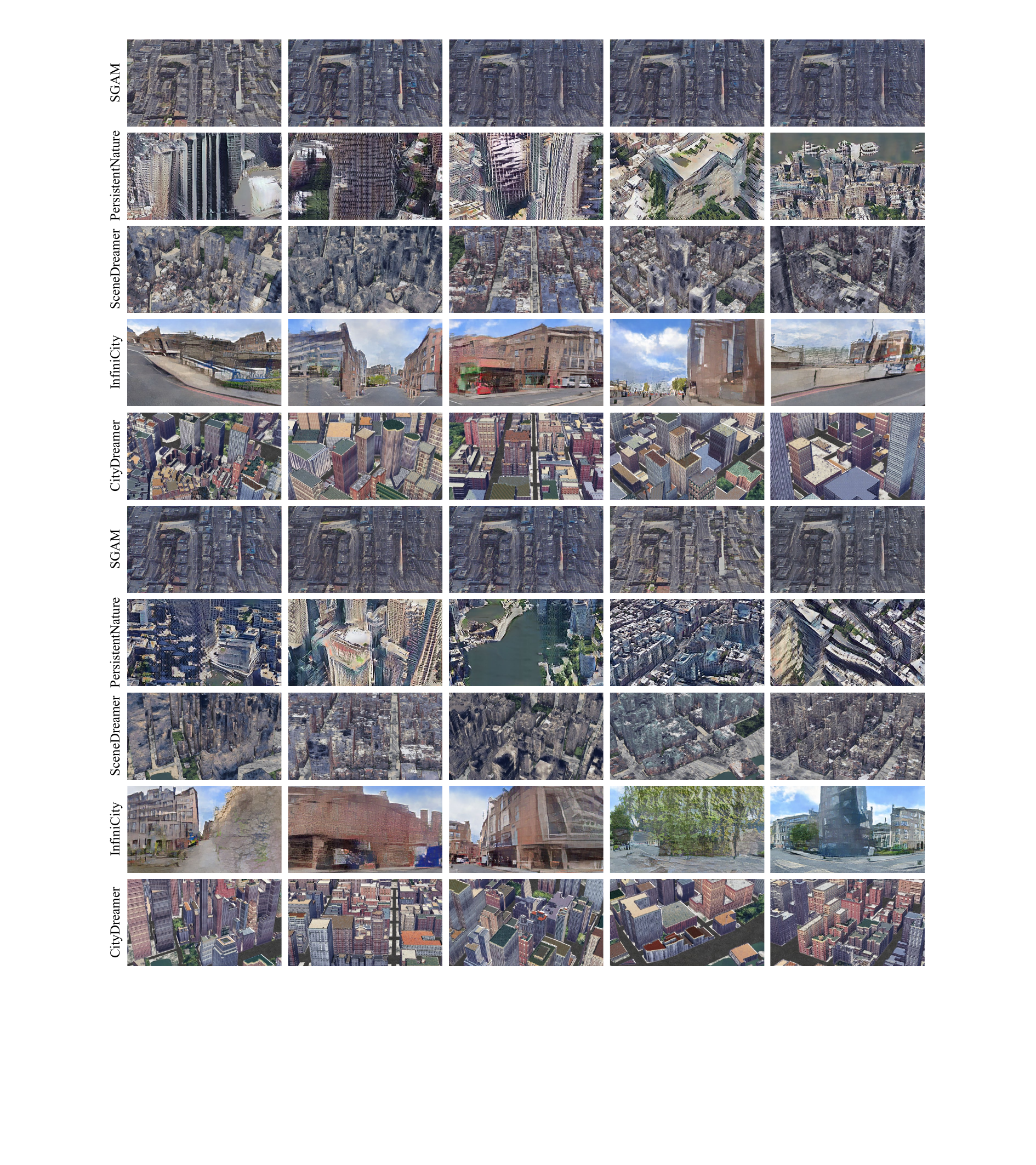}
  }
  \vspace{-6 mm}
  \caption{\textbf{Qualitative comparison.} The proposed CityDreamer produces more realistic and diverse results compared to all baselines. Note that the visual results of InfiniCity~\cite{DBLP:conf/iccv/LinLMCSYT23} are provided by the authors and zoomed for optimal viewing.}
  \label{fig:citygen-comparison-more}
  \vspace{-10 mm}
\end{figure}

In Figure~\ref{fig:citygen-comparison-more}, we provide more visual comparisons with state-of-the-art methods.
We also encourage readers to explore more video results available in the project page.

\clearpage
\section{Interactive Demo}
\label{sec:interactive-demo}

\begin{figure}
  \centering
  \resizebox{\linewidth}{!}{
    \includegraphics{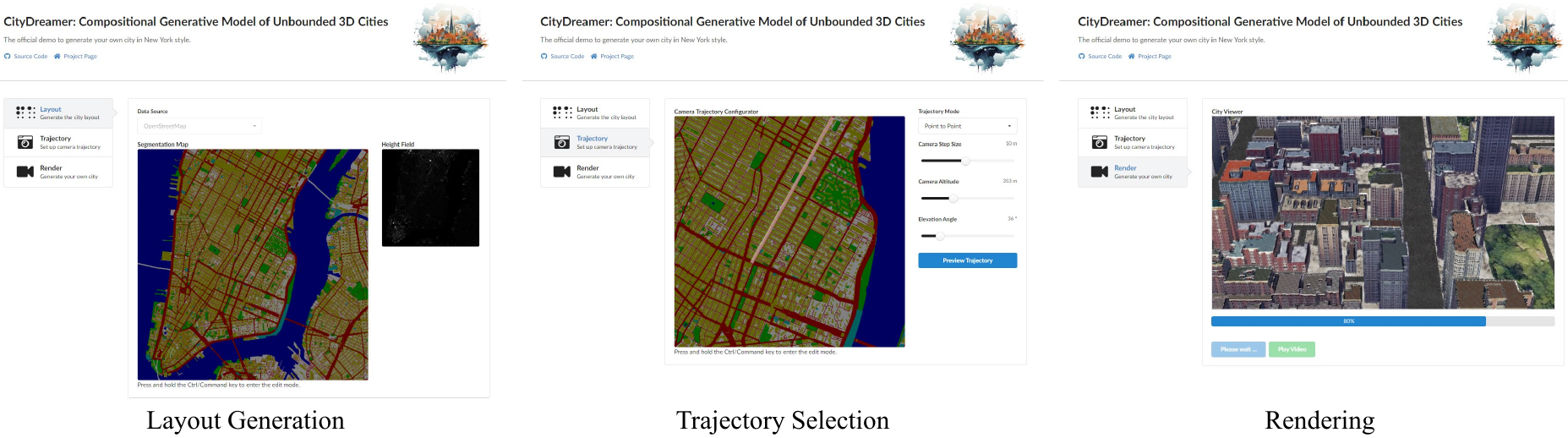}
  }
  \caption{\textbf{The screenshots of the interactive demo.} This interactive demo allows users to create their own cities in an engaging and interactive manner. We encourage the readers to explore the video demo available in the project page.}
  \label{fig:interactive-demo}
\end{figure}

We develop a web demo that allows users to interactively create their own cities.
The process involves three main steps: layout generation, trajectory selection, and rendering, as illustrated in Figure~\ref{fig:interactive-demo}. 
Users can manipulate these steps to create customized 3D city scenes according to their preferences.

During the layout generation phase, users have the option to create a city layout of arbitrary sizes using the unbounded layout generator, or they can utilize the rasterized data from OpenStreetMap directly.
This flexibility allows users to choose between generating layouts from scratch or using existing map data as a starting point for their 3D city.
Additionally, after generating the layout, users can draw masks on the canvas and regenerate the layout specifically for the masked regions.

During the trajectory selection phase, users can draw camera trajectories on the canvas and customize camera step size, view angles, and altitudes. 
There are three types of camera trajectories available: orbit, point to point, and multiple keypoints. 
Once selected, the camera trajectory can be previewed based on the generated city layout, allowing users to visualize how the city will look from different perspectives before finalizing their choices.

Finally, the cities can be rendered and stylized based on the provided city layout and camera trajectories.

\end{document}